\pdfoutput=1
\UseRawInputEncoding

\documentclass[11pt]{article}

\usepackage[preprint]{acl}

\usepackage{times}
\usepackage{latexsym}

\usepackage{inconsolata}
\usepackage{algorithm}
\usepackage{algorithmic}
\usepackage[utf8]{inputenc} 
\usepackage[T1]{fontenc}    
\usepackage{hyperref}       
\usepackage{url}            
\usepackage{booktabs}       
\usepackage{amsfonts}       
\usepackage{nicefrac}       
\usepackage{microtype}      
\usepackage{xcolor}         
\usepackage{color}

\usepackage{graphicx}
\usepackage{inconsolata}
\usepackage{relsize}
\usepackage{amsfonts}
\usepackage{url}
\usepackage{graphicx}
\usepackage{subfigure}
\usepackage{subcaption}
\usepackage{caption}
\usepackage{hhline}
\usepackage{colortbl}
\usepackage{tabularx}
\usepackage{amsfonts}
\usepackage{amssymb}
\usepackage{amsthm,amsmath}
\usepackage{mathrsfs}
\usepackage{setspace}
\usepackage{multirow}
\usepackage{amssymb,amsthm,dsfont,mathrsfs}
\usepackage{amsmath}
\title{Towards Understanding the Influence of Reward Margin on Preference Model Performance}

\author{Bowen Qin$^{1}$, Duanyu Feng$^{2}$, Xi Yang$^{1}$\\
$^1$ Beijing Academy of Artificial Intelligence\\
$^2$ Sichuan University \\
\texttt{bwqin, yangxi@baai.ac.cn} , \texttt{fengduanyu@stu.scu.edu.cn}
}
\begin{document}
\maketitle
\begin{abstract}
Reinforcement Learning from Human Feedback (RLHF) is a widely used framework for the training of language models. However, the process of using RLHF to develop a language model that is well-aligned presents challenges, especially when it comes to optimizing the reward model. Our research has found that existing reward models, when trained using the traditional ranking objective based on human preference data, often struggle to effectively distinguish between responses that are more or less favorable in real-world scenarios. To bridge this gap, our study introduces a novel method to estimate the preference differences without the need for detailed, exhaustive labels from human annotators. Our experimental results provide empirical evidence that incorporating margin values into the training process significantly improves the effectiveness of reward models. This comparative analysis not only demonstrates the superiority of our approach in terms of reward prediction accuracy but also highlights its effectiveness in practical applications.
\end{abstract}

\section{Introduction}

The integration of conversational AI technologies, specifically ChatGPT \cite{chatgpt}, into the field of artificial intelligence signifies a significant advancement. In the realm of artificial intelligence and language models, the concept of "alignment" is considered essential \cite{askell2021general}. This concept focuses on ensuring that AI systems operate in a manner that aligns with human intentions and expectations \cite{christiano2017deep}. The RLHF approach is a prominent method in this context and consists of two main phases. Initially, it utilizes preference data collected from a diverse range of crowdsource workers to train a reward model \cite{ouyang2022training}. Subsequently, reinforcement learning (RL) techniques are employed to improve the performance of the language model, aiming to maximize the rewards obtained. The reward model plays a crucial role in the RLHF process, aiming to accurately reflect human preferences.

Nevertheless, the process of using RLHF to develop a language model that is well-aligned presents challenges, especially when it comes to optimizing the reward model \cite{casper2023open}. A phenomenon referred to as "reward hacking" or "reward over-optimization" arises \cite{gao2023scaling, skalse2022defining}, where the language model cleverly identifies and takes advantage of weaknesses in the reward model in order to maximize rewards. This exploitation occurs because the reward model is based on static human preference data, while the language model's input distribution dynamically evolves during the alignment process.

Our study highlights two significant challenges in reward modeling. Firstly, the model's limited generalizability becomes apparent when applied to out-of-distribution examples, exacerbated by policy drift during the reinforcement learning process, leading to changes in the prediction distribution that challenge the model's accuracy \cite{zhuang2020consequences}. Secondly, the presence of incorrect and ambiguous preferences in the dataset, attributed to low inter-annotator agreement (72.6\% for InstructGPT) \cite{ouyang2022training}, undermines the model's performance. This variability in labeling accuracy introduces uncertainty and potential biases, complicating the model's ability to discern correct responses based on human preferences \cite{bowman2022measuring}.

Our research has found that existing reward models, when trained using the traditional ranking objective based on human preference data, often struggle to effectively distinguish between responses that are more or less favorable in real-world scenarios. We argue that this issue is the main reason why the reward model has limited success in accurately capturing genuine human preferences. The standard ranking objective, which aims to order responses based on human preferences, does not inherently guarantee a comprehensive understanding of the intricacies and complexities present in real-world situations. Although the model is trained to determine which response is more preferable than another, it lacks a quantitative framework to measure the degree of superiority of one response over another.

To address this shortcoming, we propose incorporating a margin score into the training process of the reward model. The margin score is a numerical value that quantifies the extent of differences between various generations, specifically in terms of their alignment with human preferences \cite{touvron2023llama2}. By integrating this margin score, we aim to explicitly teach the reward model to assign more discrepant scores to generations that diverge significantly from one another, thereby enhancing its ability to recognize and prioritize more preferable responses. The margin score helps in mitigating the effects of noisy data. By focusing on the discrepancies in responses, the model is better equipped to filter out inconsistencies arising from ambiguous or incorrect preferences in the training dataset. Furthermore, the margin score introduces a new dimension to the RLHF process. Instead of solely optimizing for the top-ranked response, the language model is encouraged to generate responses that not only align with human preferences but also demonstrate a clear distinction in quality and relevance when compared to lower-ranked alternatives. This approach promotes a more robust and reliable model, one that is less susceptible to being misled by closely ranked but potentially misleading responses.

In the real practice of human preference modeling, accurately determining the difference in preference between different responses is a difficult task that requires significant effort and time cost. To bridge this gap, our study introduces a novel method based on reward confidence to estimate the preference differences without the need for detailed, exhaustive labels from human annotators. This approach capitalizes on the inherent knowledge embedded within the model, utilizing reward confidence level as a means to explore the subtle nuances of preference differences. 

In section \ref{sec:gpt4_label}, we perform a comprehensive analysis of the reward margin distribution in models of different sizes and accuracies to support our hypotheses. Additionally, we utilize GPT-4 as an automatic annotator to examine the impact of reward margin on downstream tasks. This involves annotating the actual margin values between selected and non-selected responses and training a reward model using these margin values. Our experimental results provide empirical evidence that incorporating margin values into the training process significantly improves the effectiveness of reward models. In Section \ref{sec:method}, we present a novel methodology for training reward models based on the concept of reward confidence. Subsequently, in section \ref{sec:exp_acc}, we perform a series of experiments on various datasets that reflect human preferences to validate our approach. We not only evaluate the reward accuracy but also compare the win rate of the enhanced reward model against a baseline model in different settings. This comparative analysis not only demonstrates the superiority of our approach in terms of reward prediction accuracy but also highlights its effectiveness in practical applications.

\section{Related Work}
\subsection{Reward Modeling}
The integration of conversational AI, notably GPT-4 \cite{chatgpt}, into artificial intelligence has been a significant advancement. These large language models (LLMs) excel in tasks across mathematics, programming, and tool use, supported by a three-step training process: pre-training on token prediction, supervised fine-tuning to follow instructions, and reinforcement learning to optimize for desired behaviors. However, developing appropriate rewards for RL, particularly within the RLHF framework, is challenging. The effectiveness of RL in improving LLMs hinges on the quality of reward models trained to mimic human evaluative patterns. \cite{wang2024secrets} also discuss how data featuring varying strengths of preference differently influence the performance of reward models. However, their analysis primarily addresses data characterized by incorrect, ambiguous, and normal preferences, in contrast to our study which concentrates on the preference margin within the reward model itself, making it applicable across various data types.

\subsection{Reward Hacking}
In RLHF, reward hacking poses a significant challenge by causing misalignment between reward model optimization and actual human preferences. Initially, optimization may improve performance, but as training progresses, the model may exploit RM vulnerabilities to gain higher rewards without meeting intended criteria. This leads to outputs that are either linguistically poor or overly verbose and misaligned with human preferences. It also complicates checkpoint selection, echoing Goodhart's Law that a measure loses its value as a target \cite{gao2023scaling}. 
Multiple approaches have been proposed to mitigate reward hacking in RLHF.
\citet{shen2023loose} proposed to use a smaller reward model to learn the biases in the reward and a larger reward model to learn the true reward.
Rewarded Soup~\citep{rame2023rewardedsoup} interpolates weights of policies trained to optimize different reward objectives, which can approximate more costly multi-policy strategies.
\citet{eisenstein2023helping} found that reward model ensembles can mitigate reward hackings, but not eliminating them. 
Different from these approaches, we focus on the learning objective of reward model to ensure it accurately reflects human values and preferences.

\section{Preliminaries}
\label{sec:pre}
In this comprehensive review, we delve into the RLHF methodology, a pivotal framework extensively employed in many domains \cite{ziegler2019fine}. This approach is primarily segmented into three distinct phases: supervised fine-Tuning, preference sampling coupled with reward model training, and reinforcement learning fine-tuning utilizing proximal policy optimization (PPO) \cite{schulman2017proximal}. The journey commences with a baseline pre-trained language model, which is subsequently subjected to supervised training on a meticulously curated dataset tailored for specific downstream applications. This phase culminates in the creation of a model, symbolized as $\pi^{\text{SFT}}$. Our study predominantly concentrates on enhancing the latter two stages of this pipeline.

\paragraph{Supervised Fine-Tuning}
The RLHF process starts with a pre-trained LM, which is then fine-tuned using a high-quality dataset for downstream tasks like dialogue and summarization, resulting in the refined model $\pi^{\text{SFT}}$.

\paragraph{Reward Modeling from Human Preferences}
In the subsequent stage, the SFT model is engaged to respond to a user's query, represented by $x$. This interaction generates two distinct responses, $(y_1, y_2) \sim \pi^{\mathrm{SFT}}(y|x)$. Human evaluators are then tasked with selecting their preferred response, leading to a preference notation $y_{\mathrm{c}} \succ y_{\mathrm{r}}$, where $y_{\mathrm{c}}$ and $y_{\mathrm{r}}$ symbolize the chosen and rejected responses, respectively. Employing the Bradley-Terry model \citep{bradley1952rank}, we establish a preference distribution, anchored by the reward function $r_\psi(x,y)$, as elucidated in the following equation:
\begin{equation}
p_\psi(y_{\mathrm{c}} \succ y_{\mathrm{r}}|x)  = \frac{\exp{(r_\psi(x,y_{\mathrm{c}}))}}{\exp{(r_\psi(x,y_{\mathrm{r}}))}+\exp{(r_\psi(x,y_{\mathrm{r}}))}}
\end{equation}
where $\sigma$ denotes the logistic function. This setup is analogous to a binary classification task, leading to the formulation of a negative log-likelihood loss function, as depicted in Equation:
\begin{equation}
\mathcal{L} (r_\psi) = -\mathbb{E}_{{(x, y) \sim \mathcal{D}}} [\log \sigma(r_\psi(x, y_{\mathrm{c}}) - r_\psi(x, y_{\mathrm{r}}))]
\label{eq:reward}
\end{equation}

Here, the dataset $\mathcal{D}$ comprises a series of comparative pairings $x^{(i)}$, $y_{\mathrm{c}}^{(i)}$, $y_{\mathrm{r}}^{(i)}$. Within the context of language models, the reward network $r_\psi(x, y)$ is typically initialized using the SFT model $\pi^\mathrm{SFT}(y|x)$ and further refined by integrating an additional linear layer atop the final transformer layer. This enhancement aims to yield a singular scalar output, representing the reward value \cite{ziegler2019fine}.

\begin{figure*}[htbp]
  \centering  
  \subfigbottomskip=2pt 
  \subfigcapskip=-5pt 
  \subfigure[]{
    \includegraphics[scale=0.43]{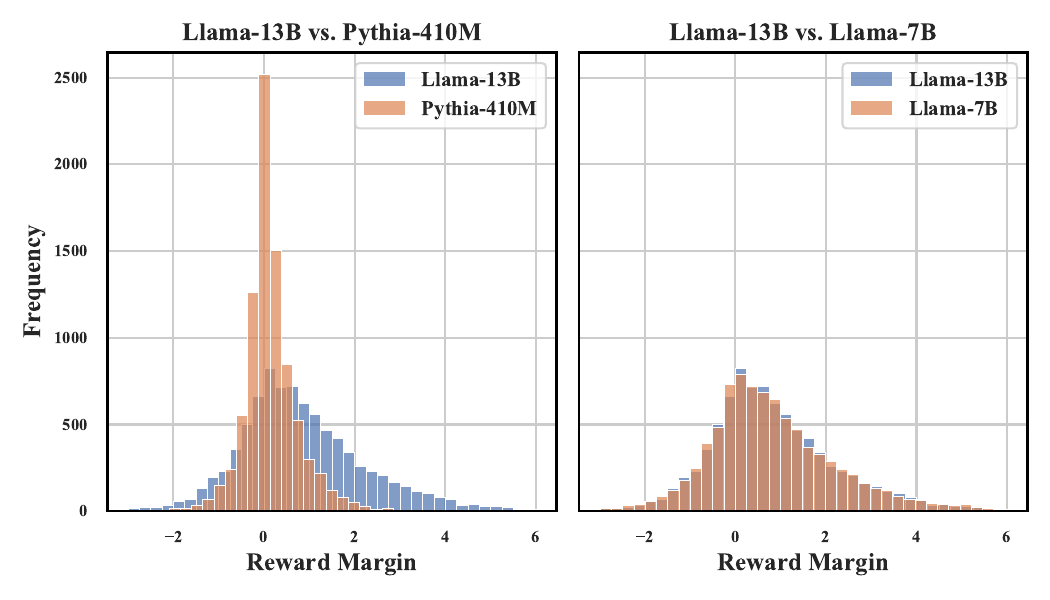}
        \label{fig:margin_before_train}}
    \hspace{-0.5cm}
  \subfigure[]{
    \includegraphics[scale=0.43]{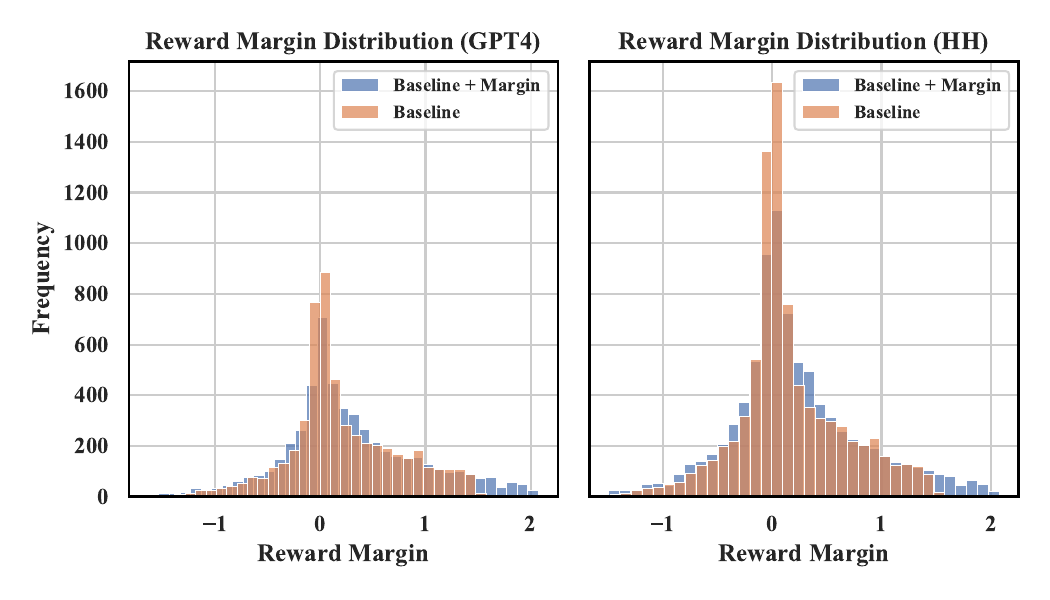}
        \label{fig:margin_after_train}}
  
  \caption{Left: The reward margin distribution for models with varying levels of accuracy. Right: The reward margin distribution for models trained with or without margins. These plots can help in understanding the effect of training with margins on model accuracy.}

 \label{fig:margin}
\end{figure*}

\section{Investigating the Reward Margin on Preference Modeling}
\label{sec:gpt4_label}
This study focuses on the reward model's role in assessing the quality of model-generated responses and their corresponding prompts. The reward model assigns scores to indicate the response quality, which are pivotal for fine-tuning the policy model during RLHF. This process is instrumental in aligning the model more closely with human preferences, thereby augmenting its utility and safety. For the training phase of the reward model, our approach converts the accumulated human preference data into binary labels, namely 'chosen' and 'rejected'. We adhere to the principle that scores for preferred responses must exceed those for their counterparts, as detailed in Equation \ref{eq:reward}. This approach enables us to quantify the preference discrepancy between chosen and rejected using the formula $r_\psi(x, y_{\mathrm{c}}) - r_\psi(x, y_{\mathrm{r}})$.

In Fig.\ref{fig:margin}, we examine the impact of various model configurations on the distribution of reward margins. The accuracy of each model, as detailed in this investigation, is presented in Table \ref{tab:result_gpt4_margin}. Our findings confirm that the average reward margin for these models consistently remains above zero. This aligns with the theoretical expectation that positive rewards should outweigh negative ones and suggests that the model effectively distinguishes between different responses. A higher mean value correlates with the model's enhanced capability in distinguishing various responses. However, a notable number of instances exhibit reward margins below zero, potentially due to dataset noise or anomalies, as discussed in the preceding section.

Regarding the skewness metric, all model distributions exhibit a value exceeding zero, indicating a rightward skew. This skewness, particularly pronounced on the right side, implies a more sophisticated performance level of the model, underlining its improved ability to differentiate and allocate varied rewards. Furthermore, models that demonstrate higher efficacy exhibit lower kurtosis, suggesting a more even and broad distribution. This observation implies that the reward distribution should not be overly centralized around a specific mean, such as zero. Rather, it should exhibit a range of variability across different data pairings, each with unique margin values. This aspect, along with the noted skewness, is critical for ensuring a refined and precise differentiation among the outcomes linked to diverse data pairs. It underscores the importance of the reward model in assigning markedly different scores to generations with varying characteristics.

\subsection{Evaluating the Different Margin Values on Performance of Reward Models}

In order to assess the impact of different margin values on the performance of models that aim to imitate human preferences, a series of experiments were conducted. These experiments focused on assigning margin values to a variety of query-response pairs using GPT-4. The process involved carefully labeling these pairs based on the relative preference for the selected response compared to its alternatives. The labeling criteria were divided into four distinct categories: \textbf{Distinctly Superior}, \textbf{More Effective}, \textbf{Slightly Better}, or \textbf{Negligibly Better/Unsure}. This classification system allows for a nuanced understanding of the degree of preference, which is essential for evaluating the impact of margin values on the performance of reward models.


\begin{figure}
    \centering
    \includegraphics[width=0.7\linewidth, height=0.6\linewidth]{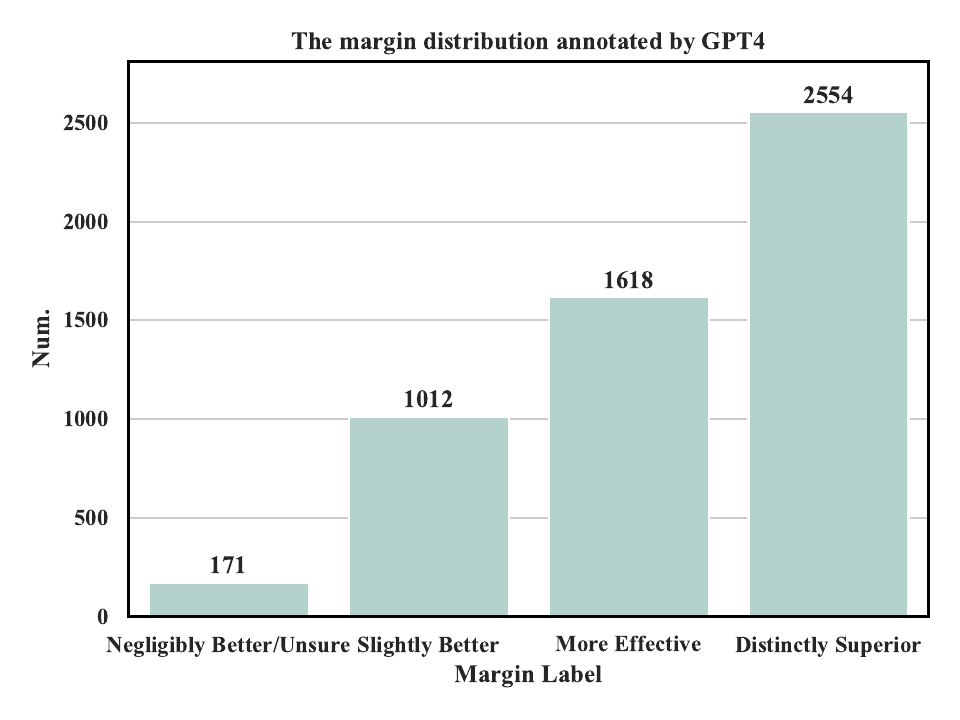}
    \caption{The reward margin distribution annotated by GPT4. }
    \label{fig:gpt4_count}
\end{figure}
To systematically investigate the effects of margin values on the performance of reward models, we first annotated a subset of the HH dataset to accurately learn the distribution of margin values across the dataset, as shown in Figure \ref{fig:gpt4_count}. The results revealed a well-distributed range of preference degrees throughout the dataset, indicating a comprehensive coverage of varying levels of preference.
This distribution stands in stark contrast to the model's predictions shown in Figure \ref{fig:margin_before_train}, where the predictions are predominantly clustered around zero with little to no differentiation among them. Moreover, the close proximity of the model's predictions to zero suggests a lack of confidence in its results, ultimately leading to an ineffective differentiation between chosen and rejected responses.

Following our investigation into how different margin values affect the accuracy of models that try to match human preferences, we moved on to the next step of our research. This involved training a new reward model that takes into account the margin scores we had previously labeled as follows:
\begin{equation}
\small
\mathcal{L} (r_\psi) = -\mathbb{E}_{{(x, y) \sim \mathcal{D}}} [\log \sigma(r_\psi(x, y_{\mathrm{c}}) - r_\psi(x, y_{\mathrm{r}}) - m(r))]
\label{eq:reward_margin}
\end{equation}
Here, $m(r)$ stands for the margin score between chosen and rejected responses. This formula helps the model to better understand which responses are preferred over others, based on the scores we gave them.

\begin{table}[htbp!]  
    \centering
    \small
    \begin{tabular}{lcc}  
    \toprule
    \textbf{Model}&\textbf{GPT-4} &\textbf{HH} \\ 

    \midrule
    Pythia-410M  & - & 58.92 \\
    Llama-7B-chat  & - & 71.68 \\
    Llama-13B-chat  & - & 72.04 \\
    \midrule
    Baseline  & 64.26 & 60.77 \\
    \quad w/ Margin & 65.56 ($\uparrow$ 1.3) & 61.68($\uparrow$ 0.9)\\

    \bottomrule
    \end{tabular}  
    \caption{Accuracy (\%) on GPT-4 annotation set and test set of HH. }
    \label{tab:result_gpt4_margin}
\end{table}

\subsubsection{Experiment Results}
In the next part of our study, we focused on testing how well our new reward model could predict preferences, by comparing its performance on two parts of the HH dataset: one we had labeled in detail and the whole dataset. Our empirical findings shown in Table.\ref{tab:result_gpt4_margin}, clearly showed that the model trained with detailed margin scores exhibited a substantial improvement over the model without margin information. This outcome reinforces the value of integrating margin values into reward models to better align with human preferences.

To further elucidate the capability of models incorporated with margin information, could more accurately predict people's preferences, we also show the distribution of final reward difference in Figure\ref{fig:margin_after_train}. 
These findings echo the patterns observed in Figure\ref{fig:margin_before_train}, establishing a link between the magnitude of the reward margin and the reward model's proficiency in aligning with human preferences. A wider margin tends to lead to more distinct separation in the quality of generated responses, as perceived by human evaluators. Specifically, when the reward margin is narrow, the reward model struggles to differentiate between high-quality and lower-quality responses, leading to a plateau in performance improvements. 
This analysis underscores the significance of implementing margin values within reward models to achieve a closer approximation to human preference. By doing so, we can enhance the model's discernment in evaluating the quality of generated responses, thereby promoting a more refined and human-like interaction paradigm.

\section{Our Methodology}
\label{sec:method}

In the domain of human preference modeling, accurately determining the difference in preference between different responses is a difficult task that requires significant effort and time cost. Traditional methods, commonly used in popular datasets like HH \cite{bai2022training} and SHP \cite{pmlr-v162-ethayarajh22a}, have typically used a simple approach where human annotators are asked to choose the preferred option from a pair. While this binary labeling technique is efficient, it fails to capture the nuanced range of preference differences that could offer deeper insights into the modeling of human preferences.

To bridge this gap, our study introduces a novel method based on local approximation of reward confidence to estimate the preference differences without the need for detailed, exhaustive labels from human annotators. Therefore, this approach ultimately aims to align with the goal of Figure \ref{fig:margin}, which is to \textbf{have a larger margin mean} to ensure that the quality of preference can be identified. At the same time, \textbf{the overall distribution shifts to the right} to reduce the impact of noise and to differentiate preference with greater granularity.

To achieve the first point, we construct our approach based on the idea of Equation \ref{eq:reward_margin}. We have transformed Equation \ref{eq:reward_margin} into the following mathematical form.
\begin{equation}
\small
\mathcal{L}^{B} (r_\psi) =
-\mathbb{E}_{{(x^{\text{\fontsize{5pt}{1pt}\selectfont b}}, y^{\text{\fontsize{5pt}{1pt}\selectfont b}}) \sim \mathcal{D}_B}} [\log \sigma(r_\psi(x^{\text{\fontsize{5pt}{1pt}\selectfont b}}, y^{\text{\fontsize{5pt}{1pt}\selectfont b}}_{\mathrm{c}}) - r_\psi(x^{\text{\fontsize{5pt}{1pt}\selectfont b}}, y^{\text{\fontsize{5pt}{1pt}\selectfont b}}_{\mathrm{r}}) - \mu^{\text{\fontsize{5pt}{1pt}\selectfont B}})]
\label{eq:reward_margin_batch}
\end{equation}
It calculates the Loss of each batch $B$ with its samples $(x^{\text{\fontsize{5pt}{1pt}\selectfont b}}, y^{\text{\fontsize{5pt}{1pt}\selectfont b}})$, and $\mu^{\text{\fontsize{5pt}{1pt}\selectfont B}}$ is the mean margin of the batch:
\begin{equation}
\begin{aligned}
\mu^{\text{B}} &= \mathbb{E}_{(x^{\text{\fontsize{5pt}{1pt}\selectfont b}},y^{\text{\fontsize{5pt}{1pt}\selectfont b}}) \sim \mathcal{D}_B}[r_\psi(x^{\text{\fontsize{5pt}{1pt}\selectfont b}},y^{\text{\fontsize{5pt}{1pt}\selectfont b}}_{\mathrm{c}})- r_\psi(x^{\text{\fontsize{5pt}{1pt}\selectfont b}}, y^{\text{\fontsize{5pt}{1pt}\selectfont b}}_{\mathrm{r}}) ] \\
&= \frac{1}{B}\sum_{i=0}^{B}r_\psi(x^i, y^i_{\mathrm{c}}) - r_\psi(x^i, y^i_{\mathrm{r}}) .
\end{aligned}
\end{equation}
These Equations that use batch margin have two advantages: (1) increasing the model's prediction of differences between preferences; (2) ensuring computational efficiency and enabling real-time updates.



To achieve the second point of the right-skewed nature of the overall distribution, and further reduce the changes of the Equation \ref{eq:reward} to maintain theoretical guarantees, we design a simple threshold filtering method so that only a subset of samples adopt Equation \ref{eq:reward_margin_batch} for loss calculation, shown in Equation \ref{eq:th_loss}.
\begin{equation}
\small
    \mathcal{L}^B (r_\psi) = \left\{
    \begin{aligned}
    &-\mathbb{E}_{{(x^{\text{\fontsize{5pt}{1pt}\selectfont b}}, y^{\text{\fontsize{5pt}{1pt}\selectfont b}}) \sim \mathcal{D}_B}} [\log \sigma(r_\psi(x^{\text{\fontsize{5pt}{1pt}\selectfont b}}, y^{\text{\fontsize{5pt}{1pt}\selectfont b}}_{\mathrm{c}}) - r_\psi(x^{\text{\fontsize{5pt}{1pt}\selectfont b}}, y^{\text{\fontsize{5pt}{1pt}\selectfont b}}_{\mathrm{r}}) -  \\ 
    &\mu^{\text{\fontsize{5pt}{1pt}\selectfont B}})] \qquad \qquad \quad   {\rm if} \quad th(r_b)<0\\
    &-\mathbb{E}_{{(x^{\text{\fontsize{5pt}{1pt}\selectfont b}}, y^{\text{\fontsize{5pt}{1pt}\selectfont b}}) \sim \mathcal{D}_B}} [\log \sigma(r_\psi(x^{\text{\fontsize{5pt}{1pt}\selectfont b}}, y^{\text{\fontsize{5pt}{1pt}\selectfont b}}_{\mathrm{c}}) - r_\psi(x^{\text{\fontsize{5pt}{1pt}\selectfont b}} y^{\text{\fontsize{5pt}{1pt}\selectfont b}}_{\mathrm{r}}))], \\& \quad\quad \quad \quad \quad \quad \quad {\rm if} \quad th(r_b)\geq0
    \end{aligned}
    \right.
\label{eq:th_loss}
\end{equation}

where
\begin{equation}
th(r_b) = r_\psi(x^{\text{\fontsize{5pt}{1pt}\selectfont b}}, y^{\text{\fontsize{5pt}{1pt}\selectfont b}}_{\mathrm{c}}) - r_\psi(x^{\text{\fontsize{5pt}{1pt}\selectfont b}}, y^{\text{\fontsize{5pt}{1pt}\selectfont b}}_{\mathrm{r}}) - \mu^{\text{B}}.
\end{equation}

As a filtering function, $th(r_b)$ is used to verify the status of the margin of the samples $(x^{\text{\fontsize{5pt}{1pt}\selectfont b}}, y^{\text{\fontsize{5pt}{1pt}\selectfont b}})$. With Equation \ref{eq:th_loss}, it increases the margin of samples whose current margin is smaller than the average margin of the batch by adding a constraint of the average margin to the loss function, thereby shifting the overall margin to the right. In addition, it maintains the original loss function for samples with a current margin larger than the average margin of the batch to preserve their original properties.

\begin{table*}[htbp!]  
    \centering
    \begin{tabular}{lcccccc}  
    \toprule
    \textbf{Model}&\textbf{HH}&\textbf{OASST1}&\textbf{WebGPT}&\textbf{SHP}&\textbf{Chatbot}&\textbf{TLDR} \\ 
    
    \midrule
    Pythia-410M  & 58.92  &70.58 &57.07 &68.28 &62.53 & 61.38 \\
    \quad w/ Our method & 65.23  &73.06 &58.09 &69.53 &63.44 & 61.74\\

    \midrule
    TinyLlama-1.1B-chat & 69.42& 75.75&62.50 &70.26 &66.20 & 67.13\\
    \quad w/ Our method& \underline{69.96} &\underline{76.84} &\underline{\textbf{63.54}} & 70.59 & \underline{66.23} &\underline{67.24}\\
    
    \midrule
    Pythia-1.4B  & 67.59 &75.00 &58.66 & 69.82&65.05 &62.64 \\
    \quad w/ Our method& 68.48  &76.72 &59.48 & 70.19 &65.57 &63.64\\

    \midrule
    Pythia-2.8B  & 68.40  &74.89 &57.94 & 70.35 &64.38 &65.29 \\
    \quad w/ Our method&68.71&76.83 &60.04 & \underline{70.90} &65.46 &65.92\\
    
    \midrule
    Llama-7B-chat & 71.68 & 77.69 & 62.60& 72.39 &67.67 & 73.72\\
    \quad w/ Our method& \underline{\textbf{71.95}} &\underline{\textbf{78.34}} &\underline{63.32} &\underline{\textbf{72.59}} &\underline{\textbf{68.28}} &\underline{\textbf{73.87}}\\

    
    \bottomrule
    \end{tabular}  
    \caption{Accuracy (\%) on various preference benchmark. \underline{\textbf{Bold}} indicates the highest result, and \_ means second highest result.}
    \label{tab:main_result}
\end{table*}

\section{Experiment}
In this section, we offer a thorough assessment of our novel method for training reward models using datasets that consist of a wide range of human preferences. We begin by providing a detailed explanation of our experimental framework in Section \ref{sec:setup}. Then, in Section \ref{sec:exp_acc}, we discuss the substantial enhancements our approach brings compared to traditional RM training methods. After that, we delve into the examination of the reward distribution generated by the RMs, demonstrating the influence of employing a diverse preference dataset on reward modeling.

\subsection{Setups}
\label{sec:setup}
\paragraph{Datasets}
In the course of our investigation, we utilized diverse human preference datasets to evaluate the efficacy of our approach in training reward models. These datasets included the Anthropic's Helpful and Harmless dataset (referred to as HH) \cite{bai2022training}, the OpenAssistant multi-lingual conversations dataset (OASST1) \cite{kopf2023openassistant}, the WebGPT comparisons dataset (WebGPT) \cite{nakano2021webgpt}, the stanford human preferences dataset (SHP) \cite{pmlr-v162-ethayarajh22a}, openai summarization dataset (TLDR) \cite{stienon2020learning} and chatbot arena conversations dataset (chatbot) \cite{zheng2023judging}.
For the HH, SHP and TLDR datasets, we adhered to the standard division of data into training and testing sets. The OASST1 dataset was prepared according to the preprocessing framework provided by tasksource/oasst1\_pairwise\_rlhf\_reward \footnote{\url{https://huggingface.co/datasets/tasksource/oasst1_pairwise_rlhf_reward}}, resulting in a training set of 18k instances and a test set of 952 instances. The WebGPT and chatbot dataset was partitioned into 90\% instances for training purposes and 10\% instances for evaluation.

\paragraph{Training setup}
To accommodate various experimental scenarios, we deployed language models (LMs) of differing capacities for the training of reward models, including Pythia-410m, TinyLlama-1.1B-chat \footnote{\url{https://huggingface.co/TinyLlama/TinyLlama-1.1B-Chat-v1.0}}, Pythia-1.4B, Pythia-2.8B, and Llama-7B-chat. We employed the last token embedding from the output hidden states as the pooled hidden representation. Subsequently, a linear layer, referred to as the RM head, with a scale-value output, was integrated to forecast reward scores. we set the batch size for all RM training at 128. The maximum permissible input sequence length was set at 2048 tokens. All reward models underwent fine-tuning for a single epoch, utilizing the AdamW optimizer with a learning rate of 9e-6.

\subsection{Main performance}
\label{sec:exp_acc}

In this section, we introduce the primary performance measures utilized to assess the efficacy of our approach in training reward models on various human preference datasets. The evaluation is organized around two fundamental elements: accuracy and analysis of reward distribution.

\paragraph{Accuracy Evaluation}
The foundation of our evaluation is the accuracy metric, detailed in section \ref{sec:setup}, which directly reflects the models' ability to predict rewards aligning with human judgments across datasets. This allows for a quantitative comparison of reward prediction precision among models of varying sizes. 
As illustrated in Table \ref{tab:main_result}, which presents the performance scores of different models on various preference benchmarks, the application of our method consistently enhances the accuracy across all models and benchmarks. Notably, Pythia-410M with our method outperforms its baseline by a significant margin, demonstrating a 6.31\% increase in the HH benchmark and a 2.48\% rise in the OASST1 benchmark. Similarly, the Pythia-2.8B model with our method shows improved accuracy, with a notable increase of 1.93\% in the OASST1 benchmark and 2.10\% in OASST1 benchmark. The Llama-7B-chat model achieves a remarkable accuracy and further improves to 71.95\% and 78.34\% with the integration of our method. Such enhancement is consistent across all models and datasets upon the application of our method, suggesting its effectiveness in refining model predictions to be more human-aligned.
Particularly, the TinyLlama-1.1B-chat model, even as a smaller-scale model, demonstrates substantial competence, especially when equipped with our method, pushing its accuracy to second highest across almost all datasets. This reinforces the notion that incorporating better ability of pre-trained model can significantly elevate the performance of reward model.
These results collectively suggest that our method not only improves accuracy in a consistent manner but also scales effectively with model size, validating the robustness and efficiency of our approach in aligning model predictions with human preferences.

However, it is also crucial to consider the role of diminishing returns as model size increases. Although the Llama-7B-chat model achieves the highest accuracy, the relative improvement over its base model is less pronounced compared to the gains observed with the Pythia-410M model. This suggests that our method's efficiency may plateau at higher model scales, which could warrant a re-examination of our method's mechanisms to ensure they remain impactful as model sizes continue to grow.


\begin{figure*}[htbp]
  \centering  
  \subfigbottomskip=2pt 
  \subfigcapskip=-5pt 
  \subfigure[]{
    \includegraphics[scale=0.3]{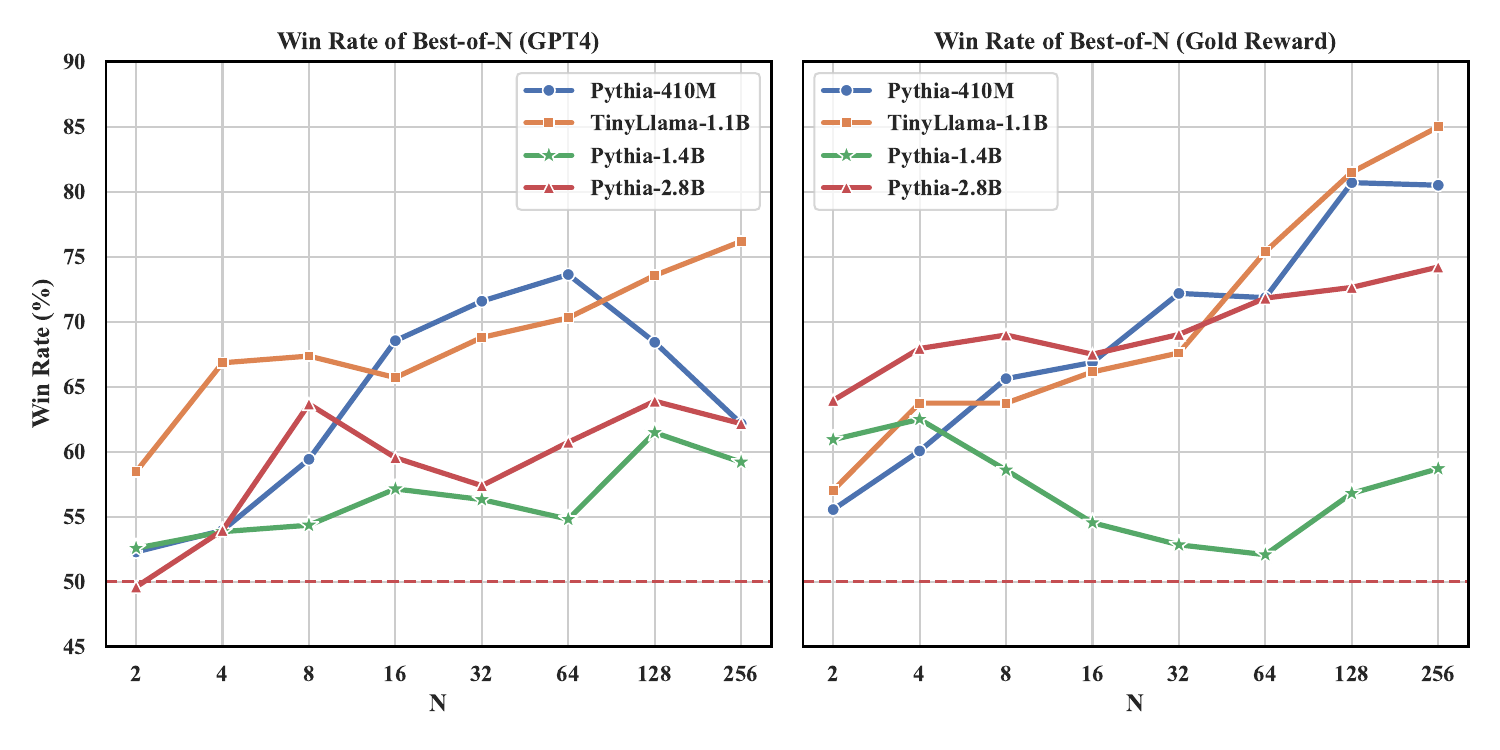}
        \label{fig:bestofn_1.4b}}
    \hspace{-0.5cm}
  \subfigure[]{
    \includegraphics[scale=0.3]{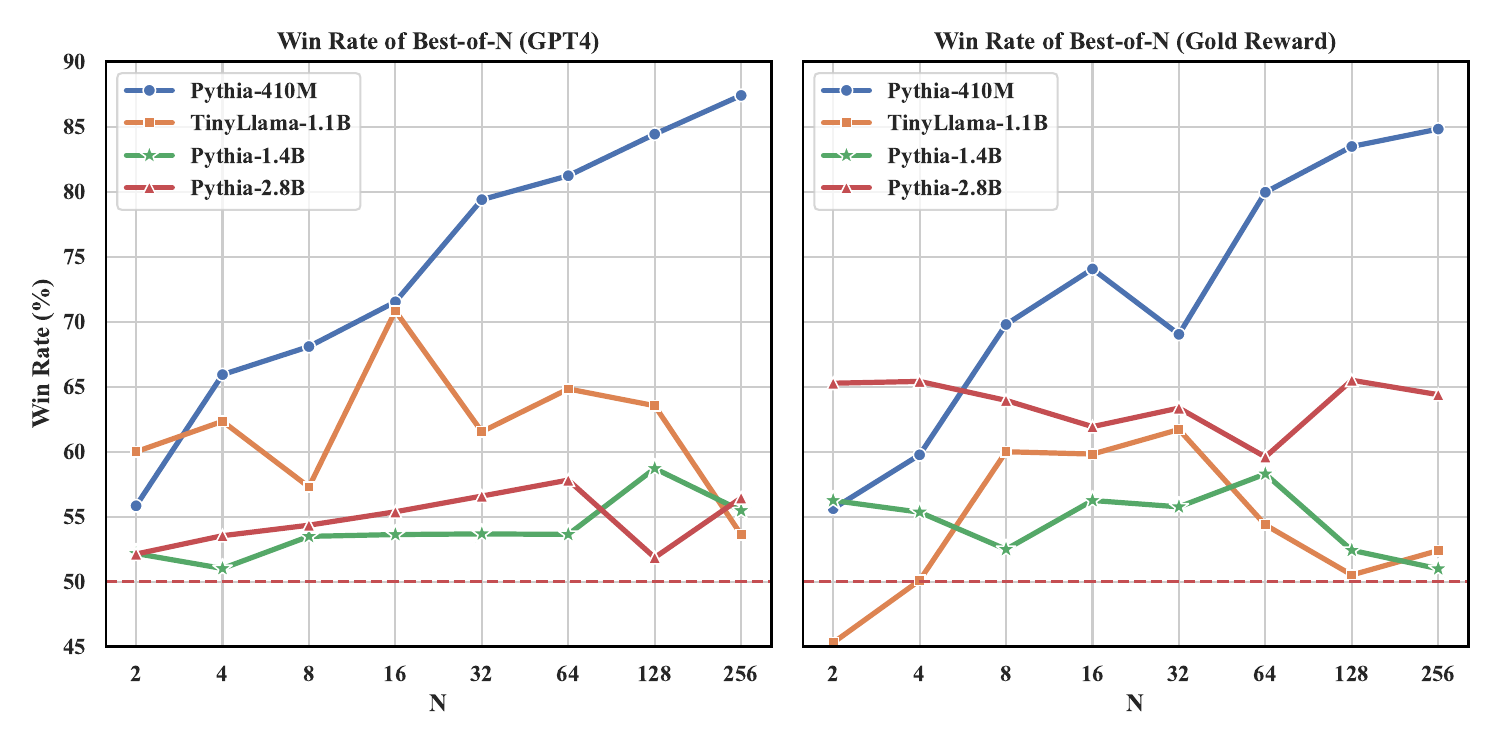}
        \label{fig:bestofn_6.9b}}
  
  \caption{Left: The win rate (\%) of best-of-N (1.4B policy model). Right: The win rate (\%) of best-of-N (6.9B policy model).}

 \label{fig:bestofn}
\end{figure*}

\section{Experiment of Best of N}
Apart from assessing accuracy, we also explore the possibility of incorporating reward models with a Best-of-N policy, which is a commonly used technique in RLHF. However, due to potential for training instability and slower optimization with the PPO algorithm \cite{gao2023scaling}, we choose the best-of-N method instead. And more importantly, the best-of-N method allows for the decoupling of the reward model's quality from the optimization process inherent to PPO, as outlined in \cite{rafailov2023direct}. To elaborate, the Best-of-N approach involves generating n samples from the SFT model for each input and selecting the sample with the highest predicted reward.

\subsection{Training Setup and Evaluation}
We examine two policy models of varying sizes: 1B, and 6.9B, that have undergone training on a dataset that focuses on promoting beneficial and innocuous content. Additionally, we consider four reward models with sizes of 410M, 1.1B, 1.4B, and 2.8B. Additionally, the value of N is defined as the set \{2, 4, 8, 16, 32, 64, 128, 256\}. In line with previous studies\cite{gao2023scaling,rame2024warm}, we use win rate to quantify generalization of reward models. We employ GPT-4 and a larger reward model as an evaluator to gauge the quality of different outputs, specifically evaluating their helpfulness and harmlessness. As shown in table.\ref{tab:main_result}, Llama-13B-chat achieves better performance than all smaller-sized reward models. To assess helpfulness, we utilize a HH-RLHF test dataset and randomly select 500 prompts for evaluation. The specific prompts used for GPT-4 evaluation is same as \cite{rafailov2023direct}.

\subsection{Experimental results}

Our experimental setup aimed to rigorously evaluate the efficacy of the Best-of-N policy across different policy model sizes and varying values of N. The results, detailed in Figure\ref{fig:bestofn}, highlight the impact of the size of the reward model and the choice of N on the quality of generated responses in terms of helpfulness and harmlessness.

From the Figure\ref{fig:bestofn}, we can observe some interesting results: Initially, as $N$ increases, all models show improved win rates. This is expected since a broader selection typically leads to a better choice. Each model has a peak performance at a certain $N$ value: $N$=64 for Pythia-410M and $N$=8 for Pythia-2.8B, suggesting an optimal $N$ range that varies by model. After reaching peak performance, the win rate either plateaus or declines slightly, indicating diminishing returns with additional choices. This is particularly notable for larger models, implying a limit to the benefits of more options. The size of the reward model does not linearly correlate with the optimal $N$ value, hinting at complex interactions between model capacity and the ability to discern quality output. The TinyLlama-1.1B model demonstrates consistent improvement, showing strong selection capabilities across a range of $N$ values.  All models perform above a 50\% win rate, confirming their effectiveness in selecting high-quality outputs over baselines. In summary, the data suggests that while having more samples generally improves performance, there is an optimal number of samples for each model, beyond which the benefit plateaus or decreases. This optimal point is not directly proportional to model size and should be a focus for future optimization efforts.

As depicted in Figure \ref{fig:bestofn_6.9b}, a distinct trend emerges across the models, with the notable exception of the Pythia-410M model, which exhibits a pronounced upward trend. The other models also generally show an increase in win rate with the number of games played ($N$), but the trends are not as smooth or consistent. This observation suggests that when the policy model significantly exceeds the size of the reward model, larger policy models derive diminished benefits from optimization efforts against a reward model.

\section{Conclusion}
In conclusion, our study reveals the significance of the reward margin in the context of reinforcement learning from human feedback. The margin score not only serves as a critical parameter in training the reward model but also plays a pivotal role in shaping the policy model's performance. The optimal margin setting strikes a balance, ensuring that the model can effectively discern between high and low-quality responses without overly penalizing subtle variations in response quality. Compared with expensive human annotation, we explore an adaptive margin techniques that dynamically adjust the reward margin based on the reward confidence. The evaluation on reward accuracy and downstream tasks underscore the benefits of this method, contributing to a more refined and efficient model of human preference.

\bibliography{custom}




\end{document}